\documentclass{article}

\usepackage[preprint]{eiml_style_2025}
\usepackage{graphicx} 
\usepackage{xcolor}
\usepackage{amsmath}
\usepackage{amsfonts}       
\usepackage{algorithm}
\usepackage{algorithmic}
\usepackage{graphicx}    
\usepackage{afterpage}   
\usepackage{lipsum}      
\usepackage{url}
\usepackage{hyperref} 
\usepackage{enumitem}
\usepackage{subcaption}
\DeclareMathOperator{\given}{|}

\title{Where to Measure: Epistemic Uncertainty-Based Sensor Placement with ConvCNPs}

\author{%
    Feyza Eksen\\
    Marine Data Science\\
    University of Rostock\\
    \texttt{feyza.eksen@uni-rostock.de}
    \And
    Stefan Oehmcke\\
    Visual and Analytic Computing in Ocean Technologies\\
    University of Rostock\\
    \texttt{stefan.oehmcke@uni-rostock.de}
    \AND
    Stefan Lüdtke\\
    Marine Data Science\\
    University of Rostock\\
    \texttt{stefan.luedtke@uni-rostock.de}
}

\begin{document}

\maketitle

\begin{abstract}
    Accurate sensor placement is critical for modeling spatio-temporal systems such as environmental and climate processes. Neural Processes (NPs), particularly  Convolutional Conditional Neural Processes (ConvCNPs), provide scalable probabilistic models with uncertainty estimates, making them well-suited for data-driven sensor placement. However, existing approaches rely on total predictive uncertainty, which conflates epistemic and aleatoric components, that may lead to suboptimal sensor selection in ambiguous regions. To address this, we propose expected reduction in epistemic uncertainty as a new acquisition function for sensor placement. To enable this, we extend ConvCNPs with a Mixture Density Networks (MDNs) output head for epistemic uncertainty estimation. Preliminary results suggest that epistemic uncertainty driven sensor placement more effectively reduces model error than approaches based on overall uncertainty.
\end{abstract}


\section{Introduction}

\afterpage{%
    \clearpage
    \begin{figure}[t]
        \centering
        \includegraphics[width=\textwidth]{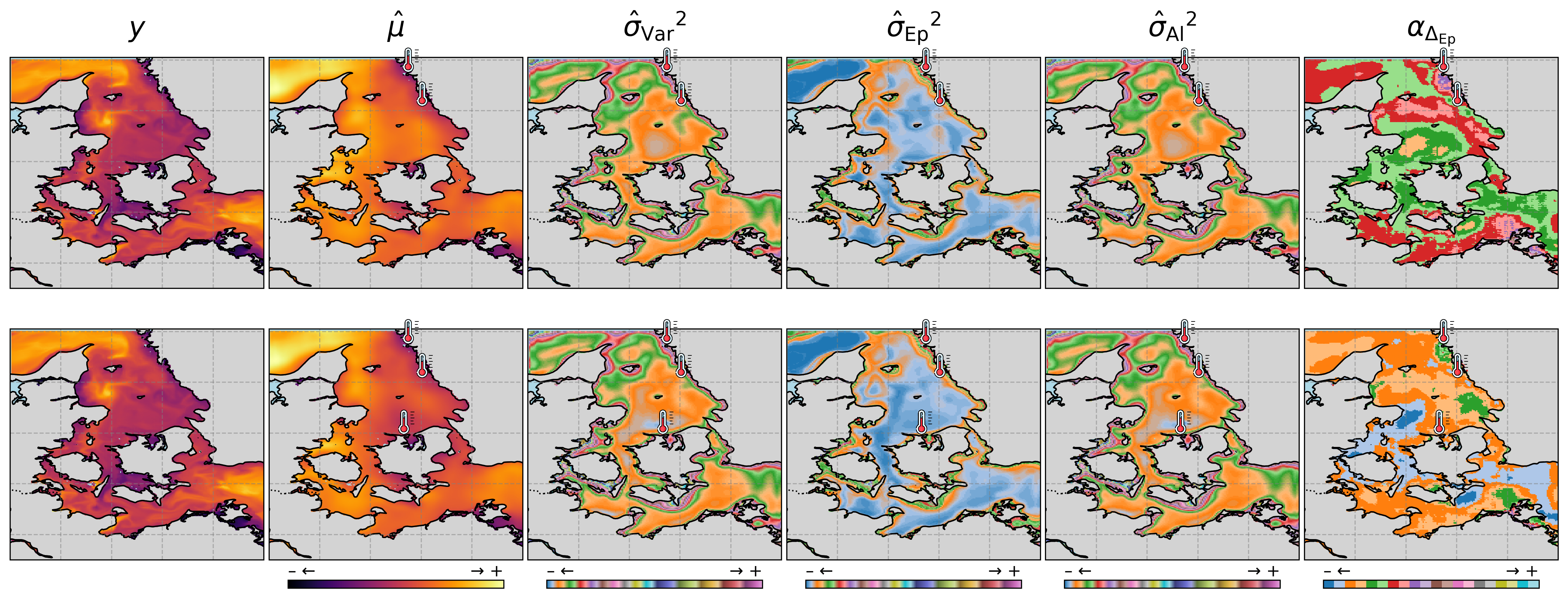}
        \caption{
            \textbf{Greedy sensor placement in the Western Baltic Sea region using proposed Epistemic-Uncertainty-based acquisition function.}
            The leftmost panel shows the true \textit{Sea Surface Temperature} $y$ on January 1, 2022. The top row presents the ConvCNP model’s prior estimates conditioned on two sensor observations, while the bottom row shows the updated posterior after placing a third sensor at the optimal location $\arg\min \alpha_{\Delta_\mathrm{Ep}}$ (indicated in the top-right panel). The predictive mean $\hat{\mu}$ becomes more accurate and the model error is reduced after conditioning on the additional sensor. The quantities ${\hat{\sigma}_\mathrm{Var}}^2$, ${\hat{\sigma}_\mathrm{Ep}}^2$, and ${\hat{\sigma}_\mathrm{Al}}^2$ denote the estimated predictive, epistemic, and aleatoric variance, respectively.
        }
        \label{fig:main_figure}
    \end{figure}
}

Selecting optimal locations for environmental sensors is a central problem in many scientific and engineering domains, including environmental monitoring, climate and weather forecasting, air quality assessment, and oceanographic observation. Given a limited sensor budget, the objective is to select sensor locations that best characterize a high-dimensional system by maximizing information gain or minimizing predictive uncertainty. This sensor placement problem has been widely studied under both optimization-based and probabilistic frameworks~\cite{krause2008nearoptimalsensor, brunton2022datadrivenscience}.

Recent progress in data-driven modeling has introduced Neural Processes (NPs) as flexible probabilistic models that learn distributions over functions, offering scalable alternatives to Gaussian Processes. The Convolutional Conditional Neural Processes (ConvCNPs)~\cite{gordon2019convcnp} represent a notable member of the Neural Process family~\cite{dubois2020npf, garnelo2018neural, garnelo2018conditional}, combining translation equivariance with efficient uncertainty estimation and has shown promise for spatio-temporal tasks relevant to sensor placement~\cite{andersson2023deepsensor}.

Sensor placement is guided by acquisition functions, which describe the utility of obtaining measurements at a certain location. 
Existing approaches often derive these functions from uncertainty estimates, such as those produced by ConvCNPs~\cite{andersson2023deepsensor}. 
However, these estimates conflate epistemic (model) and aleatoric (data) uncertainty. This can lead to suboptimal sensor placements, especially in ambiguous regions where accurate estimation of epistemic uncertainty would provide the most informative guidance.

To address this limitation, we explore an extension of ConvCNPs with Mixture Density Networks (MDNs)~\cite{bishop1994mdn} output layer to enable a decomposition of predictive uncertainty. 
Building on prior works that utilized MDNs for epistemic uncertainty estimation~\cite{choi2018uncertaintyaware, choi2021activelearning}, we investigate how epistemic uncertainty can be estimated within ConvCNP model framework and leveraged as an acquisition signal for sensor placement (see Figure~\ref{fig:main_figure}). Unlike typical active learning setups where ground truth is queried iteratively, our sensor placement setting assumes that measurements become available only after all sensors are deployed. 
In this ongoing work, our preliminary experiments suggest that proposed epistemic-driven sensor placement may more effectively reduce model error.

Our contributions are as follows:
\begin{itemize}[itemsep=0.6pt, topsep=0.5pt, leftmargin=3em]
    \item We extend ConvCNPs with an MDN-based output for richer uncertainty modeling.
    \item We suggest an estimator for epistemic uncertainty within ConvCNP model framework.
    \item We present preliminary results indicating that the expected reduction in epistemic uncertainty can serve as an effective acquisition function for sensor placement.
\end{itemize}
\section{Preliminaries}

In the following, we present the specific type of Neural Process employed in this work, introduce the sensor placement problem, and describe acquisition functions used in prior work.

\subsection{Convolutional Conditional Neural Processes}

Convolutional Conditional Neural Processes (ConvCNPs) can be regarded as a flexible, learned analogue of Gaussian Processes (GPs), in which both the mean and covariance functions are parameterized by neural networks rather than fixed kernels. In contrast to typical neural network architectures, they can handle non-gridded data, which makes them suitable for sensor placement tasks~\cite{andersson2023deepsensor}. Formally, given a context set \(\mathcal{C} = \{(x_i, y_i)\}_{i=1}^{N_c}\), where \(x_i \in \mathbb{R}^d\) and \(y_i \in \mathbb{R}\), the predictive distribution over target variables \(y\) at target locations \(x\) conditioned on the given context set is \(p(y \given x, \mathcal{C})\). A task \(\mathcal{D}\) is defined as \((\mathcal{C}, \mathcal{T})\) where \(\mathcal{T}\) is a target set \(\mathcal{T} = \{(x_j, y_j)\}_{j=1}^{N_t}\).

The predictive marginal is modeled as a univariate Gaussian distribution \(p(y \given x, \mathcal{C}) = \mathcal{N}(y; \mu(x, \mathcal{C}), \sigma^2(x, \mathcal{C}))\), where the parameters $\mu$ and $\sigma^2$ depend on both $x$ and $\mathcal{C}$ as follows: The model first projects the context set \(\mathcal{C}\) onto a uniform grid through a localized kernel function via \textit{\text{SetConv($\mathcal{C}$)}},
which is then processed by convolutional layers such as \textit{\text{R=U-Net(SetConv($\mathcal{C}$))}}~\cite{andersson2023deepsensor}. When queried with a (potentially off-grid) target location \(x\), it interpolates the feature map \textit{\text{R}} at $x$. The interpolated value is passed to an MLP to compute the predictive distribution parameters \(\mu, \sigma^2\). Network parameters are optimized via gradient descent to minimize the negative log-likelihood (NLL) over target locations.

\subsection{Sensor Placement \& Acquisition Functions}

The sensor placement problem can be formulated as a subset selection task. Let the set of candidate sensor locations be $\mathcal{L} \subset \mathbb{R}^d$ with finite  number of locations, and the objective is to select a subset $S \subseteq \mathcal{L}$ that maximizes an objective function $F(S)$:

\begin{equation*}
    S^\star = \arg\max_{S \subseteq \mathcal{L}} F(S) \quad \text{s.t. } |S^\star| = N_s,
\end{equation*}
The objective \(F(S)\) (e.g., likelihood or negative RMSE on the target set $\mathcal{T}$) can usually not be evaluated during sensor placement, as target labels are not available. Thus, an \emph{acquisition function} $\alpha(x)$ is used as a proxy. It quantifies the expected utility of placing a sensor at a location $x$,  capturing how informative the resulting observations are about the underlying spatial field. 

When historical data capturing the system dynamics are available, probabilistic models can be trained to act as surrogate models of the underlying system. ConvCNPs, in particular, have been shown to be effective for this task~\cite{andersson2023deepsensor}. Their predictive variance \(\sigma^2(x, \mathcal{C})\) can be used as a basis for defining the acquisition function, guiding the placement of sensors to regions where additional observations will most effectively improve the model. 
Specifically, the expected reduction in mean predictive variance over the target locations, after appending $x$ to the context set $\mathcal{C}$ has been proposed as an acquisition function $\alpha_{\Delta_\text{Var}}$~\cite{andersson2023deepsensor} (see Supplementary Material Section \ref{subsubsec:aqfun} for a formal definition).
 
Due to the combinatorial explosion in the number of subsets $S$, sensor placement is usually done via a greedy method, iteratively selecting sensor locations to add to $S$.
Following this established strategy~\cite{andersson2023deepsensor}, the greedy sensor placement procedure is (see Supplementary Material Section \ref{subsubsec:greedy_sensor} for more details):  
\begin{enumerate}
    \item Predict the mean $\hat{y}_i$ at all candidate locations $x_i \in \mathcal{L}$ by using the prior \(p_\theta(y \given x_i, \mathcal{C})\).
    \item Temporarily add each pair $(x_i,\hat{y}_i)$ to the context set \(\mathcal{C}\) and compute the average predictive variance across target locations.
    \item Select the candidate $(x_{i^*}, \hat{y}_{i^*})$ that yields the largest reduction in average predictive variance.
    \item Update the context set \(\mathcal{C}\) with the candidate $(x_{i^*}, \hat{y}_{i^*})$ and repeat the process until the required number of sensors is selected.
\end{enumerate}

\section{Epistemic Uncertainty-Based Sensor Placement}

In this section, our main contributions are introduced: An extension of ConvCNPs to decompose its variance into epistemic and aleatoric components, as well as an acquisition function based on this decomposition.

\subsection{Disentangling Aleatoric and Epistemic Uncertainty in ConvCNPs}

The predictive uncertainty produced by ConvCNPs does not distinguish between uncertainty arising from limited knowledge of the underlying function and uncertainty arising from inherent noise in the function itself . The former is defined as \textit{epistemic uncertainty} and the latter as \textit{aleatoric uncertainty} in the literature~\cite{kendall2017uncertaintiesincomputervision}.

Thus, we extend ConvCNPs to represent (estimates of) these types of uncertainties. We integrate Mixture Density Networks (MDNs)~\cite{bishop1994mdn} with ConvCNPs, where each mixture component corresponds to a distinct output head. Similar to~\cite{choi2018uncertaintyaware, choi2021activelearning}, we characterize epistemic uncertainty as the disagreement between the mixture components and aleatoric uncertainty as the spread of each individual component, reflecting the intrinsic noise within the data. Together, these two form of uncertainty decompose and sum to the overall predictive variance of the MDN.

Intuitively, each mixture component can be interpreted as a \textit{local hypothesis} about the conditional mapping \(p(y \given x, \mathcal{C}) \). When multiple components disagree, it reflects the model's uncertainty about which hypothesis best explains the observed context i.e., \textit{epistemic uncertainty}. Rather than relying on computationally expensive approaches such as ensembles~\cite{lakshminarayanan2017ensembles} or MC-Dropout~\cite{gal2016dropout}, MDNs model the predictive density directly in a single forward pass.

Formally, consider a mixture density \( p(y \given x, \mathcal{C}) \) of two univariate Gaussian densities 
\begin{equation*}
p(y \given x, \mathcal{C})
    = \sum_{k=1}^{K=2} \pi_k(x, \mathcal{C})\,\mathcal{N}\big(y;\mu_k(x,\mathcal{C}),\sigma_k^2(x,\mathcal{C})\big),
\end{equation*}
where \( \pi_k(x,\mathcal{C}) \in [0,1] \) are the mixture weights, satisfying \( \sum_{k=1}^{K=2} \pi_k(x,\mathcal{C}) = 1 \). The component means and standard deviations are \( \mu_k(x,\mathcal{C}) \in \mathbb{R} \) and \( \sigma_k(x,\mathcal{C}) > 0 \)  respectively. 

The variance \( \sigma^2 \) and mean \( \mu \) of the mixture density \( p(y \given x, \mathcal{C}) \) are given by \cite{deisenroth2020mathforml}
\begin{align*}
     \sigma^2 & = \pi_1\, \sigma_1^2 + (1 - \pi_1)\, \sigma_2^2 
+ \pi_1\, \mu_1^2 + (1 - \pi_1)\, \mu_2^2 
- \big[ \pi_1\, \mu_1 + (1 - \pi_1)\, \mu_2 \big]^2\\
    \mu &=  \pi_1\, \mu_1 + (1 - \pi_1)\, \mu_2, 
\end{align*}
 respectively. The dependencies of all variables on \( x \) and \( \mathcal{C} \) are omitted for clarity.

The total predictive uncertainty can be separated into two components, \( \sigma^2_{\mathrm{Var}}(x, \mathcal{C}) = \sigma^2_{\mathrm{Ep}}(x, \mathcal{C}) + \sigma^2_{\mathrm{Al}}(x, \mathcal{C}) \). 
Here, \( \sigma^2_{\mathrm{Ep}}(x, \mathcal{C}) \) is the inter component disagreement and \( \sigma^2_{\mathrm{Al}}(x, \mathcal{C}) \) is the expected variance within each component. They are defined as \cite{choi2018uncertaintyaware}

\begin{align*}
    \sigma^2_{\mathrm{Ep}}(x, \mathcal{C}) & = \sum_{k=1}^{K=2} \pi_k(x, \mathcal{C})\,  \big( \mu_k(x, \mathcal{C}) - \mu(x, \mathcal{C}) \big)^2\\
        \sigma^2_{\text{Al}}(x, \mathcal{C}) & = \sum_{k=1}^{K=2} \pi_k(x, \mathcal{C})\, \sigma_k^2(x, \mathcal{C}).
\end{align*}

In ConvCNPs, the estimated epistemic uncertainty \( \sigma^2_{\text{Ep}}(x, \mathcal{C}) \) originates from the model's uncertainty over possible functions consistent with the observed context set \( \mathcal{C} \). When the context points provide sufficient information to constrain the function, the model can capture the underlying functional behavior  accurately. In such cases, different plausible hypotheses about the function tend to converge, and the epistemic uncertainty decreases. Conversely, when the context points are limited or ambiguous, the predictive distribution remains broader. In this scenario, epistemic uncertainty tends to increase, signaling that the model’s inference is less constrained. Section~\ref{sec:illustrative_example} in the Supplementary Material illustrates this behavior.


\subsection{Expected Reduction in Epistemic Uncertainty as Acquisition Function}

The key idea of our approach is to leverage epistemic uncertainty estimates in ConvCNPs as a guiding criterion for selecting new sensor locations. Epistemic uncertainty quantifies ambiguity over which underlying function the model considers plausible, given the current context points. Intuitively, by targeting regions of high epistemic variance, additional measurements can most effectively reduce this uncertainty and improve the reliability of predictions across the domain.

Formally, consider a predictive model $p(y \given x, \mathcal{C})$ conditioned on a context set $\mathcal{C}$. For a candidate sensor location $x_i$, we can approximate the expected reduction in epistemic uncertainty $\Delta_{\mathrm{Ep}}$ if a measurement would be obtained there (for details, see Supplementary Material Section~\ref{sec:sensor_placement}). Placing a sensor at a location with high epistemic variance constrains the model's posterior over functions, reducing ambiguity and increasing confidence in the inferred underlying function. Thus, proposed acquisition function $\Delta_{\mathrm{Ep}}$ aims to reduce uncertainty efficiently, taking advantage of the ConvCNP’s ability to represent epistemic uncertainty via MDNs.
\newpage
\section{Experiments \& Results}

These experiments aim to evaluate the performance of the proposed acquisition function~$\Delta_\mathrm{Ep}$, compared to the previously used $\Delta_\mathrm{Var}$. 
For training the ConvCNP model for environmental sensor placement, we use the Baltic Sea Physics Reanalysis Dataset~\cite{copernicus2025dataset}, which provides physical conditions for the entire Baltic Sea area derived from the 3D ice-ocean model NEMO as gridded data with daily, monthly, and annual means\footnote{\url{https://data.marine.copernicus.eu/product/BALTICSEA_MULTIYEAR_PHY_003_011/description}}. 
As a proof of concept, we focus on daily Sea Surface Temperature (SST) (°C) as the environmental variable in the Western Baltic Sea region, spanning longitudes 9.04°E–13.99°E and latitudes 53.59°N–57.99°N. The dataset has a spatial resolution of 2~km~$\times$~2~km and covers the period from 1~January~1993 to 31~December~2023. To the best of our knowledge, this is the first study that investigates optimal sensor placement for physical variables of the Baltic Sea.

Both ConvCNP (K=1) and ConvCNP with MDN (K=2) models share identical hyperparameters, training/validation splits, and same architectural backbone with the same number of trainable parameters except for the output head. The proposed model with K=2, introduces approximately 17k additional trainable parameters via a 6-layer fully connected neural network with 64-unit hidden layers with ReLU activations, and outputs mixture component weights. The best-performing checkpoints were selected at epoch~110 for $K=1$ and epoch~153 for $K=2$ where validation NLL began to increase, indicating overfitting. (For further training details see~\ref{sec:training_details})

The sensor placement setting is defined for the inference phase under several simplifying assumptions. All sensors are assumed to be identical, and their measurements are considered to perfectly correspond to the simulated SST data. It is further assumed that no prior sensor network exists, meaning the initial sensor set is empty, and that ground-truth (GT) data are unavailable until all sensors have been deployed.

\begin{figure}[h]
    \centering
    \includegraphics[width=0.9\textwidth, keepaspectratio]{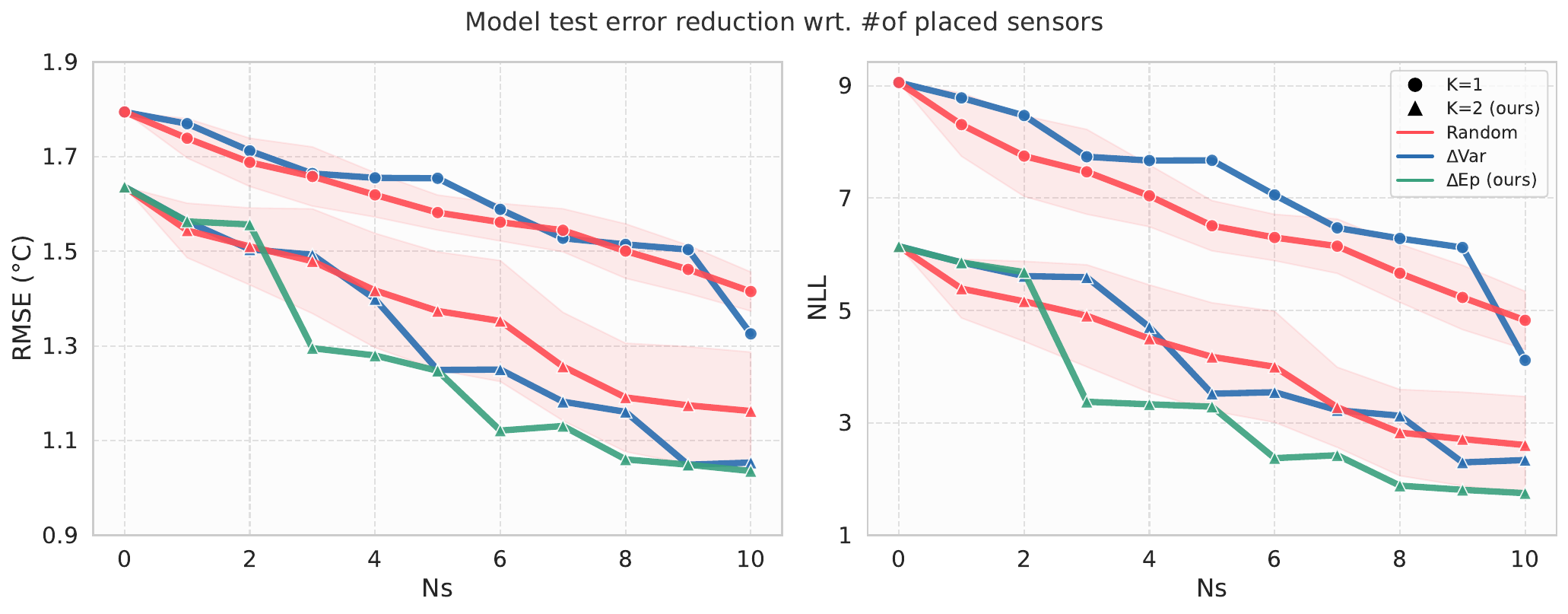}
    \caption{\textbf{Model test error RMSE (left) and negative log-likelihood (right) as a function of the number of placed sensors.} Acquisition values are computed for the test date of 1~January~2022, and the resulting sensor placements are evaluated on the same day. The random placement results are averaged over three different seeds.}
    \label{fig:2D_experiment_results}
\end{figure}

In Figure~\ref{fig:2D_experiment_results}, the proposed acquisition function $\Delta_\mathrm{Ep}$ generally achieves lower RMSE and notably lower NLL compared to $\Delta_\mathrm{Var}$. For this dataset, we observed that the estimated aleatoric uncertainty $\sigma^2_{\mathrm{Al}}$ is often significantly higher than the estimated epistemic uncertainty $\sigma^2_{\mathrm{Ep}}$. As a result, using $\Delta_{\mathrm{Var}}$ tends to favor regions with high aleatoric uncertainty rather than ambiguous regions for sensor placement. 

For $N_s = 1$, $K = 2$ both acquisition functions select the same candidate location, indicating that this particular sensor placement has the highest potential to reduce both aleatoric and epistemic uncertainty. For $N_s = 2$, the epistemic uncertainty may be underestimated when using a mixture model with $K=2$ components, which can lead to suboptimal acquisition estimates. Comparing the single-component case ($K=1$), corresponding to a standard MDN or a model without mixture components, we observe that the mixture model ($K=2$) better captures the underlying function and calibrates predictive uncertainty more effectively.

In this evaluation, the acquisition function was computed for a single test day with $N_t = 19{,}938$ and $N_l = 19{,}938$. This represents a limited evaluation, as ideally, model performance should be assessed across multiple test days. Additionally, relocating sensors daily may not be practical. A more realistic evaluation would involve averaging acquisition functions over multiple days while maintaining fixed sensor locations similar to~\cite{andersson2023deepsensor}.

Under the current assumptions, the quality of the acquisition values also depends on the accuracy of the prior point estimate $\hat{y}$, since the expected reduction is computed using the predictive mean. If prior information from station data were available, or if hyperparameter tuning improved the prior estimates, the acquisition quality would be enhanced. 
Further, for $\Delta_{\mathrm{Ep}}$, the choice of $K=2$ mixture components may not fully capture epistemic uncertainty in all regions, and increasing the number of components could further improve performance.
Despite these current limitations, these results indicate that explicitly modeling epistemic uncertainty yields more effective and informative sensor placement decisions than total uncertainty measures.
\section{Related Work}

\paragraph{Environmental Sensor Placement with ConvGNPs.} Andersson et al.~\cite{andersson2023deepsensor} proposed Convolutional Gaussian Neural Processes (ConvGNPs), which explicitly model correlations between predictions across spatial and temporal domains. The ConvGNP framework can be regarded as a flexible, learned analogue of Gaussian Processes (GPs), in which both the mean and covariance functions are parameterized by neural networks rather than relying on fixed kernels. Their work introduces a sensor placement framework~\cite{andersson2024deepsensor} that employs multiple acquisition functions for active data selection. Building upon this foundation, our approach adopts Convolutional Conditional Neural Processes (ConvCNPs), where uncertainty is modeled independently for each target point. We extend this formulation by explicitly decomposing epistemic and aleatoric uncertainties and by proposing a novel acquisition function that exploits these distinct uncertainty estimates.

\paragraph{MDNs as Epistemic Uncertainty Estimators.} Previous works~\cite{choi2018uncertaintyaware, choi2021activelearning} have explored the use of Mixture Density Networks (MDNs) for uncertainty quantification in different contexts. The former employs MDNs for imitation learning, i.e., learning from demonstrations, where quantified epistemic uncertainty is leveraged to enable efficient sampling during training. The latter integrates MDNs as the output heads for localization and classification, utilizing the combination of quantified uncertainties within an active learning framework. These approaches demonstrate the potential of MDNs as single-model methods for uncertainty quantification. In contrast, our work introduces MDNs within the framework of Neural Processes, specifically extending their application to ConvCNPs.
\section{Conclusion}

In this ongoing work, we introduced a decomposition of aleatoric and epistemic uncertainty estimates for Convolutional Conditional Neural Processes using a single forward pass, and proposed expected epistemic uncertainty reduction metric to guide sensor placement in functionally ambiguous regions. 
Preliminary experiments with 10 sensors suggest better model error reduction with the new epistemic-driven strategy.

Future work will focus on quantitative validation of the epistemic uncertainty estimates, investigating the influence of the number of mixture components, and comparing against Monte Carlo-Dropout~\cite{gal2016dropout} and ensemble baselines \cite{lakshminarayanan2017ensembles}. 
We also plan to assess how reductions in estimated epistemic uncertainty correlate with actual predictive performance (RMSE and NLL), and how relying on the predictive mean affects uncertainty estimation in successive iterations.
Further, evaluations across multiple test days are needed to assess temporal generalization (e.g., winter compared to summer performance). 
Current assumptions, such as the absence of multiple model initialization and limited hyperparameter optimization represent current limitations to be addressed in the next steps.

\newpage

\bibliographystyle{unsrt} 
\bibliography{references}

\section{Acknowledgments}
This study has been conducted using E.U. Copernicus Marine Service Information; \url{https://doi.org/10.48670/moi-00013}.
\section{Supplementary Material}

\subsection{Illustrative Example}
\label{sec:illustrative_example}

This testbed with 1D Regression Synthetic Data is designed to generate controlled datasets for analyzing the model’s ability to estimate and decompose uncertainty. Inspired by the high-noise and multi-function composition settings presented in \cite{choi2018uncertaintyaware}, we construct two corresponding scenarios: \textit{a noisy} scenario and \textit{a multiple-function} scenario.

Formally, the context set is defined as \(\mathcal{C} = \{(x_i, y_i)\}_{i=1}^{N_c}\), where \(x_i \in [-2,2]\) denotes one-dimensional spatial locations and \(y_i \in \mathbb{R}\) represents the corresponding target variable (e.g., \(y = \sin(x)\)). For training, we generate 32 tasks, each comprising a target set of \(N_t = 100\) points with uniformly sampled \(x\) target values, while the number of context points varies as \(N_c \sim [0, 5]\).

\begin{figure}[b]
    \centering
    \includegraphics[width=0.9\linewidth]{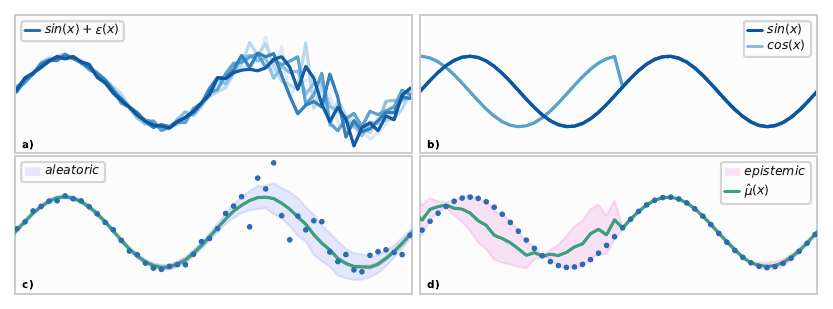}
    \caption{\textbf{Top row:} Sample training tasks for the two synthetic data distributions. \textbf{Bottom row:} Corresponding uncertainty estimates produced by the predictive model.  
    (a) Training tasks for the noisy scenario; (b) Training tasks for the multiple-function scenario;  
    (c) Aleatoric uncertainty estimated by the model \(p_\theta(y \given x, \mathcal{C})\) trained on noisy data;  
    (d) Epistemic uncertainty estimated by the model \(p_\theta(y \given x, \mathcal{C})\) trained on multiple functions.  
    In the bottom row, ground-truth target values \(y\) are shown as blue dots, and the predictive mean \( \hat{\mu} \) for target locations \(x\) is depicted as a green line for a single test task.}
    \label{fig:1D_experiments}
\end{figure}

\textbf{The noisy data.} These type of tasks are generated from a one-dimensional sinusoidal function \(y = \sin(x) + \epsilon(x) \), with \textit{Gaussian noise} added to the outputs. A Gaussian-shaped spike in noise is centered at \(x = 0.5\) with width \(0.25\) and peak magnitude \(0.5\). This creates a \textit{spatially dependent noise pattern}, where points near \(x = 0.5\) exhibit higher variance than points further away. This setup simulates regions of high measurement uncertainty and allows the model to learn input-dependent (spatial) aleatoric uncertainty.

\textbf{The multiple-function data.} These type of tasks are designed with introducing different functions in different regions of the input space. The input domain is divided into two regions. \textit{Left region} (\(x < 0\)): For each task, one of two functions is randomly selected: \(\sin(x)\) or \(\cos(x)\). This creates \textit{ambiguity} about underlying function. \textit{Right region} (\(x \geq 0\)): A fixed function \(\sin(x)\) is used for all tasks. Unlike the noisy-data case, no additional Gaussian noise is added. This dataset allows the model to learn to capture \textit{epistemic uncertainty}, i.e., uncertainty due to the unknown underlying function, particularly in the left region where multiple functions are possible.

As shown in Figure~\ref{fig:1D_experiments}, the model’s prediction of aleatoric and epistemic uncertainties aligns with the expected intuition for these isolated scenarios. In future investigations, we plan to evaluate the model on a combined dataset that integrates multiple noisy underlying functions, enabling a more comprehensive analysis of its behavior and performance.

\newpage
\subsection{Sensor Placement Details}\label{sec:sensor_placement}

\subsubsection{Model-based uncertainty reduction acquisition functions}
\label{subsubsec:aqfun}

In the following, we describe the acquisition functions in a more detail. Given context \(\mathcal{C}\) and target location \(x_j\), we denote the model's predictive variance as
\(\sigma_{\mathrm{Var},j}^2(x_j \given \mathcal{C})\) and the epistemic variance as \(\sigma_{\mathrm{Ep},j}^2(x_j \given \mathcal{C})\). For clarity, the subscript \( j \) is omitted in the following equations.

\paragraph{Expected Reduction in Mean Predictive Variance \(\Delta_\mathrm{Var}\):}
Following \cite{andersson2023deepsensor}, this acquisition function is based on the expected reduction in mean predictive variance over the target locations after appending the query sensor to the context set \(\mathcal{C}\):
\begin{align}
\alpha_{\Delta_\mathrm{Var}}(x_i) 
&= \sigma^2_{\mathrm{Var}}(x \given \mathcal{C})
- \mathbb{E}_{y_i \sim p(\cdot \given x_i, \mathcal{C})}
  \Big[
    \sigma^2_{\mathrm{Var}}(x \given \mathcal{C} \cup \{(x_i, y_i)\})
  \Big] \label{eq:expectation} \\[2mm]
&= \sigma^2_{\mathrm{Var}}(x \given \mathcal{C})
- \int p(y_i \given x_i, \mathcal{C})\,
  \sigma^2_{\mathrm{Var}}(x \given \mathcal{C} \cup \{(x_i, y_i)\})
  \, dy_i \\[2mm]
&\approx \sigma^2_{\mathrm{Var}}(x \given \mathcal{C})
- \sigma^2_{\mathrm{Var}}(x \given \mathcal{C} \cup \{(x_i, \hat{y}_i)\}) \label{eq:approx_y_mean} \\[2mm]
&\approx c
- \sigma^2_{\mathrm{Var}}(x \given \mathcal{C} \cup \{(x_i, \hat{y}_i)\}) \\[2mm]
&\approx c - \frac{1}{N_t} \sum_{j=1}^{N_t} \sigma^2_{\mathrm{Var}, j}(x_j \given \mathcal{C} \cup \{(x_i, \hat{y}_i)\})
\end{align}

As the true value \(y_i\) at the target location \(x_i\) is unknown during sensor placement, in Equation \eqref{eq:expectation} we consider the expected predictive variance  after adding the query sensor \((x_i, y_i)\) to the context. 
Computing this expectation exactly is intractable, so in Equation \eqref{eq:approx_y_mean} we approximate it by substituting the model's mean prediction \(\hat{y}_i\) at \(x_i\), following \cite{andersson2023deepsensor}.  This subtlety is crucial, as it assumes the model is reasonably accurate in predicting unobserved locations.
The predictive variance \(\sigma^2_{\mathrm{Var}}(x \given \mathcal{C})\) is treated as a constant \(c\), because it is identical for all candidate locations \(x_i\).

\paragraph{Expected Reduction in Mean Epistemic Variance \(\Delta_\mathrm{Ep}\):}

Based on decomposing the variance in aleatoric and epistemic components, we can define an epistemic uncertainty-based acquisition function. It measures the expected reduction in mean epistemic variance over targets after appending the query sensor to the context set \(\mathcal{C}\): 

\begin{align}
\alpha_{\Delta_\mathrm{Ep}}(x_i) 
&\approx c 
- \frac{1}{N_t} \sum_{j=1}^{N_t} \sigma^2_{\mathrm{Ep}, j}(x_j \given \mathcal{C} \cup \{(x_i, \hat{y}_i)\}), \label{eq:delta_ep}
\end{align}

where \(\sigma^2_{\mathrm{Ep}, j}\) is the epistemic variance at target \( x_{j} \) given the context set \(\mathcal{C} \cup \{(x_i, \hat{y}_i)\}\).

Maximizing \(\alpha_{\Delta_\mathrm{Ep}}(x_i)\) is therefore equivalent to minimizing the average epistemic variance across targets:
\begin{align*}
x_i^* = \arg\max_{x_i} \alpha_{\Delta_\mathrm{Ep}}(x_i) 
\quad \Longleftrightarrow \quad
x_i^* = \arg\min_{x_i} \frac{1}{N_t} \sum_{j=1}^{N_t} \sigma^2_{\mathrm{Ep}, j}
\end{align*}

\subsubsection{Greedy sensor placement via epistemic uncertainty reduction}
\label{subsubsec:greedy_sensor}

In Algorithm~\ref{alg:greedy_sensor}, the greedy sensor placement approach is shown in more detail. It places a fixed number of sensors at candidate locations to minimize the model's epistemic uncertainty over target locations. The algorithm is greedy, meaning that it iteratively selects the sensor that most reduces the average uncertainty across all target locations.

\begin{algorithm}[ht]
\caption{Greedy Sensor Placement Algorithm with Approximate Expected Reduction in Mean Epistemic Uncertainty}
\label{alg:greedy_sensor}
\begin{algorithmic}[1]
\REQUIRE Predictive model $p_{\theta}(y \given x, \mathcal{C})$, {\#}sensors $N_s$, {\#}target locations $N_t$, set of candidate sensor locations $\mathcal{L} = \{x_1, \dots, x_{N_l}\}$, context set $\mathcal{C}^{\star} = \emptyset$
\ENSURE Set of selected sensor locations $\mathcal{S^{\star}}$
\STATE $\mathcal{S^{\star}} \leftarrow \emptyset$
\STATE Precompute $\hat{y} \leftarrow p_{\theta}(y \given x, \mathcal{C}^{\star})$ for all $x \in \mathcal{L}$
\FOR{$n = 1$ to $N_s$}
    \STATE $\alpha^* \leftarrow \infty$, $i^* \leftarrow -1$
    \FOR{each $i \in \mathcal{L} \setminus \mathcal{S^{\star}}$}
        \STATE $\mathcal{C}_i \leftarrow \mathcal{C}^{\star} \cup \{(x_i, \hat{y}_i)\}$
        \STATE $\alpha_i \leftarrow \frac{1}{N_t} \sum_{j=1}^{N_t}
        \sigma^2_{\mathrm{Ep}, j}(x_j \given \mathcal{C}_i)$
        \IF{$\alpha_i < \alpha^*$}
            \STATE $\alpha^* \leftarrow \alpha_i$, $i^* \leftarrow i$
        \ENDIF
    \ENDFOR
    \STATE $\mathcal{S^{\star}} \leftarrow \mathcal{S^{\star}} \cup \{i^*\}$
    \STATE $\mathcal{C}^{\star} \leftarrow \mathcal{C}^{\star} \cup \{(x_{i^*}, \hat{y}_{i^*})\}$ 
\ENDFOR
\RETURN $\mathcal{S^{\star}}$
\end{algorithmic}
\end{algorithm}

The algorithm seeks to place sensors where they are most informative, using model predictions to approximate their impact on reducing epistemic uncertainty. As it allows to estimate uncertainty reduction before actual deployment, it is particularly useful in applications where sensor deployment is limited or costly, such as environmental monitoring.

\subsection{Training Details}
\label{sec:training_details}
We define the context set as $\mathcal{C} = \{(x_i, y_i)\}_{i=1}^{N_c}$, where each $x_i \in \mathbb{R}^2$ denotes a longitude–latitude coordinate and $y_i \in \mathbb{R}$ is the corresponding SST value. In addition to $\mathcal{C}$, we incorporate a set of auxiliary variables as input features, following the approach of \cite{andersson2023deepsensor}. These include (i) sea floor depth below geoid [m], which influences local SST; (ii) day of year (DOY), capturing seasonal variability; and (iii) coordinates, which help the U-Net overcome translational equivariance. The U-Net is configured with a kernel size of 5 and 64 base channels, utilizing average pooling for downsampling and bilinear upsampling for reconstruction.

We randomly sample 256 training tasks from daily data spanning the years 1993–2018, and 150 validation tasks from 2019–2021, which remain fixed during training. For each task, SST values are randomly sampled to form the context set, where the number of context points follows $N_c \sim [5, 500]$.

\subsection{Detailed Results}

\begin{figure}[ht]
    \centering
    \includegraphics[width=\textwidth, height=0.90\textheight, keepaspectratio]{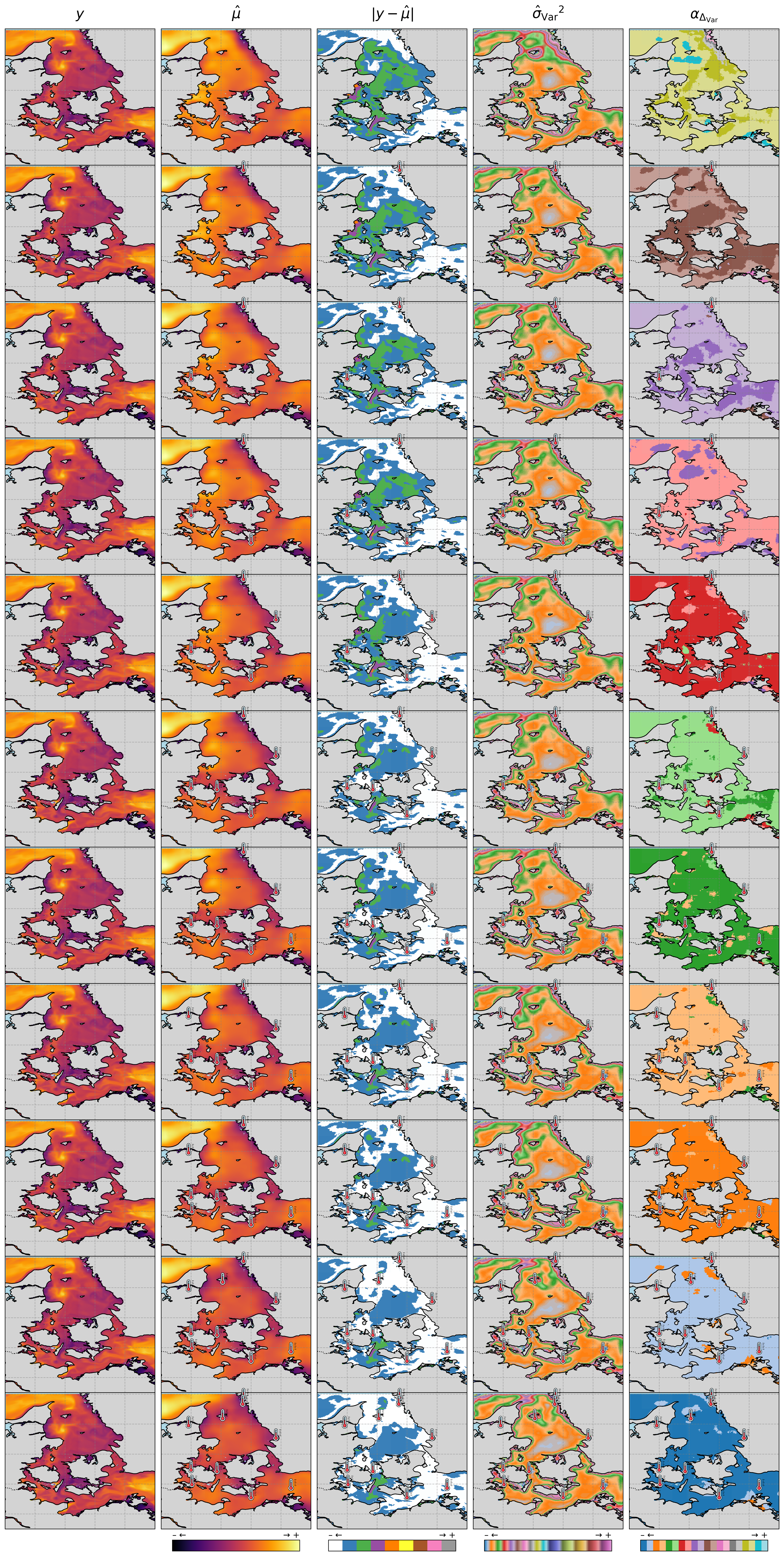}
    \caption{\textbf{Greedy sensor placement using baseline acquisition function $\alpha_{\Delta_\mathrm{Var}}$ for 10 sensors in the Western Baltic Sea region.} The leftmost panel displays the true \textit{Sea Surface Temperature} $y$ on January 1, 2022 while the rightmost panel shows acquisition values. The displayed quantities $\hat{\mu}$, $|y-\hat{\mu}|$, and ${\hat{\sigma}_\mathrm{Var}}^2$ represent the predicted mean, absolute error, and total predictive variance, respectively.}
    \label{fig:10_placement_DELTA_VAR}
\end{figure}

\begin{figure}[ht]
    \centering
    \includegraphics[width=\textwidth, height=0.90\textheight, keepaspectratio]{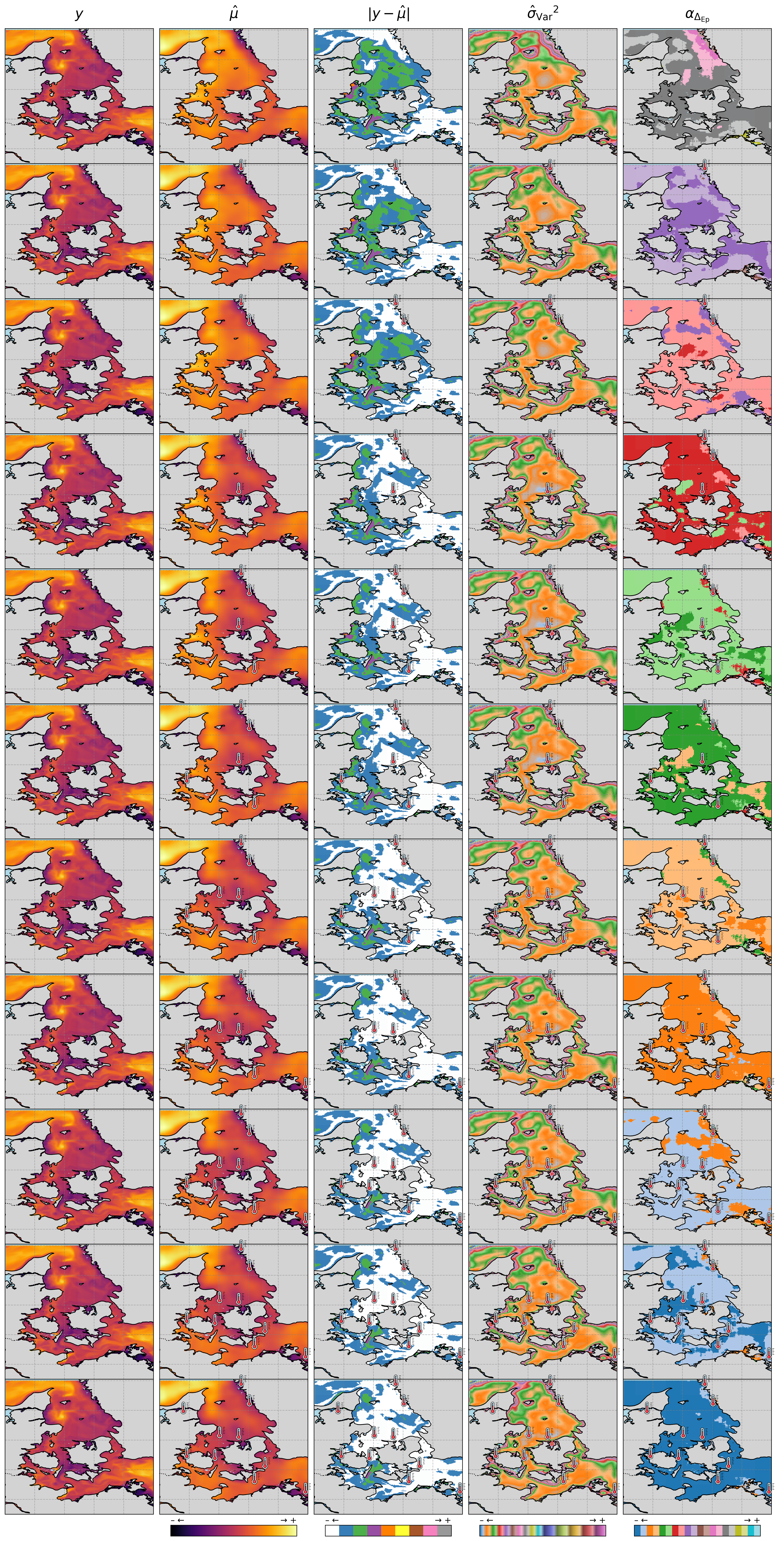}
    \caption{\textbf{Greedy sensor placement using proposed acquisition function $\alpha_{\Delta_\mathrm{Ep}}$ for 10 sensors in the Western Baltic Sea region.} The leftmost panel displays the true \textit{Sea Surface Temperature} $y$ on January 1, 2022 while the rightmost panel shows acquisition values. The displayed quantities $\hat{\mu}$, $|y-\hat{\mu}|$, and ${\hat{\sigma}_\mathrm{Var}}^2$ represent the predicted mean, absolute error, and total predictive variance, respectively.}
    \label{fig:10_placement_EPISTEMIC}
\end{figure}

This section provides a closer examination of the sensor placement results shown in Figure~\ref{fig:2D_experiment_results}. Figure~\ref{fig:10_placement_DELTA_VAR} presents the 10 greedy sensor placement results for the ConvCNP with MDN ($K=2$) using the baseline acquisition function $\alpha_{\Delta_\mathrm{Var}}$. In contrast, Figure~\ref{fig:10_placement_EPISTEMIC} shows the placements obtained with the proposed acquisition function $\alpha_{\Delta_\mathrm{Ep}}$. In both figures, each row corresponds to a data point from Figure~\ref{fig:2D_experiment_results}, evaluated conditionally on the ground truth $y$.

The absolute error $|y-\hat{\mu}|$, is visualized on an equally divided color scale, where white indicates a temperature error below $1^\circ\mathrm{C}$. Overall, the proposed acquisition function $\alpha_{\Delta_\mathrm{Ep}}$ tends to produce lower $|y-\hat{\mu}|$ values, leading to improved RMSE. Both $\arg\min \alpha_{\Delta_\mathrm{Var}}$ and $\arg\min \alpha_{\Delta_\mathrm{Ep}}$ select the same initial location, leading to identical sensor placement in the first iteration.

\end{document}